\documentclass[a4paper,conference]{IEEEtran}

\usepackage{ifpdf}

\usepackage{cite}

\ifCLASSINFOpdf
\else
\fi
\usepackage{amsmath}

\usepackage{algorithmic}

\usepackage{fixltx2e}

 \usepackage{graphicx}
\usepackage{amsmath,amssymb} % define this before the line numbering.
\usepackage{color}
\definecolor{mypink}{RGB}{218, 60, 122}

\usepackage{hyperref}
\hypersetup{
    colorlinks=true,
    urlcolor=mypink
}

\usepackage{times}
\usepackage{epsfig}

 \usepackage{transparent}
\usepackage{mathrsfs}
\usepackage{svg}
\usepackage{booktabs}

\makeatletter
\newcommand{\thickhline}{%
    \noalign {\ifnum 0=`}\fi \hrule height 1pt
    \futurelet \reserved@a \@xhline
}

\usepackage{arydshln}

\begin{document}

\title{Region-based Non-local Operation\\ 
for Video Classification}

\author{\IEEEauthorblockN{Guoxi Huang and Adrian G. Bors} \\
\IEEEauthorblockA{ Department of Computer Science,
 University of York, York YO10 5GH, UK\\
 E-mail: \{gh825, adrian.bors\}@york.ac.uk}}

\maketitle

\begin{abstract}
Convolutional Neural Networks (CNNs) model long-range dependencies by deeply stacking convolution operations with small window sizes, which makes the optimizations difficult. This paper presents region-based non-local (RNL) operations as a family of self-attention mechanisms, which can directly capture long-range dependencies without using a deep stack of local operations. Given an intermediate feature map, our method recalibrates the feature at a position by aggregating the information from the neighboring regions of all positions. By combining a channel attention module with the proposed RNL, we design an attention chain, which can be integrated into the off-the-shelf CNNs for end-to-end training. We evaluate our method on two video classification benchmarks. The experimental results of our method outperform other attention mechanisms, and we achieve state-of-the-art performance on the Something-Something V1 dataset. The code is available at: \href{https://github.com/guoxih/region-based-non-local-network}{\textit{https://github.com/guoxih/region-based-non-local-network}}.
\end{abstract}

\IEEEpeerreviewmaketitle

\section{Introduction}
\label{sec:intro}

With the rapid development of the Internet, videos have become the main multimedia resource of information, and the analysis of video information is in high demand. Video classification attracts increasing research interest, given the numerous applications for this area. As Convolutional Neural Networks (CNNs) demonstrated high capability for learning visual representations in the image domain, it is natural to attempt to apply CNNs to the video area. An effective way to extend CNN from image to video domain is by changing the convolution kernels from 2D to 3D, aka 3D CNN \cite{tran2015learning,carreira2017quo} or by adding recurrent operations to CNNs \cite{yue2015beyond,donahue2015long}.

The models based on convolutional or recurrent operations capture long-range dependencies by deeply stacking local operations with small window sizes. However, the deep stack of local operations limits the efficiency of message delivery to distant positions, and makes the optimization difficult \cite{he2016deep, hochreiter1997long}. To mitigate the optimization difficulties, Wang {\em et al.} proposed the non-local (NL) operation \cite{wang2018non} that works as a self-attention mechanism \cite{vaswani2017attention} to capture long-range dependencies directly by exploiting the inner-interactions between positions regardless of their positional distance, which we revisit in Section~\ref{sec_revisiting_nl}. However, in the NL operation, the calculation of the relation between two positions only relies on the information from these two positions while not fully utilizing the information around them. As a result, its calculation of positional relationships is not robust to noise or unrelated features, especially in high resolution, which has been emphasized in \cite{buades2005non}.  

\begin{figure*}
  \centering
  \begin{tabular}{l}
    \includegraphics[width=.99\textwidth]{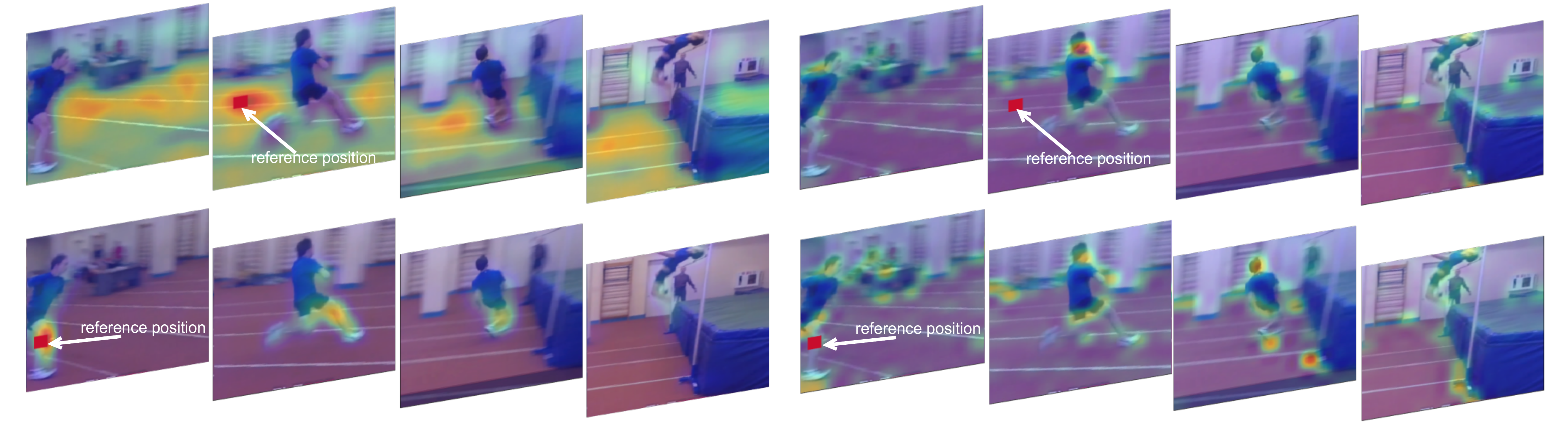}
  \end{tabular}
%   \vspace*{-0.2cm}
  (a) RNL attention maps \hspace{6cm} (b) NL attention maps
  \caption{Examples of visualizing the attention maps of RNL and NL operations in the res4 stage of ResNet on a video clip from Kinetics-400. Given a reference position, an ideal non-local operation should only highlight the regions related to the reference position. In the same video clip, the NL operation has almost the same attention maps at different reference positions while the proposed RNL operation presents query-specific attention maps, which demonstrate that the proposed RNL operation can better compute the relationships between positions.}
\label{fig:example_nl}
% \vspace*{-0.3cm}
\end{figure*}

In this paper, we investigate the non-local operation \cite{wang2018non} and propose a region-based non-local (RNL) operation based on the non-local mean concept \cite{buades2005non}, which enhances the calculation of positional relationships by fully utilizing the information from neighboring regions. The proposed RNL operation endows CNNs with a global view of input features without needing a deep stack of local operations to ease the optimization difficulties. In Figure~\ref{fig:example_nl}, we illustrate an example to demonstrate that the proposed RNL operation can better capture positional relationships than NL operation.
 There are two advantages of the proposed RNL compared with the original NL: first of all, RNL is more robust to noise or unrelated features; secondly, the RNL is more computationally efficient. Meanwhile, we present various instantiations of the RNL operation to meet different application requirements. By adding RNL operation into the off-the-shelf CNNs, we obtain a new video classification architecture named region-based non-local network. In order to evaluate the effectiveness of our method, we conduct video classification experiments on two large-scale video benchmarks, Kinetics-400 \cite{carreira2017quo} and Something-Something V1 \cite{goyal2017something}. Our models outperform the baseline and other popular attention mechanisms, and achieve state-of-the-art performance on Something-Something V1.
\section{Related Work}
% \vspace*{-0.25cm}
\label{Sec2}

% % \vspace*{+0.1cm}
\noindent \textbf{Spatio-temporal Networks.} With the tremendous success of CNNs on image classification tasks \cite{he2016deep,krizhevsky2012imagenet,Simonyan15,szegedy2015going,szegedy2016rethinking,inception_v4,xie2017aggregated,huang2017densely,tan2019efficientnet}. Some research studies have attempted to extend the applications of CNNs to video-based classification tasks \cite{simonyan2014two, yue2015beyond, carreira2017quo, li2018videolstm, huang2020learning}. Among them, the two-stream model \cite{simonyan2014two} and its variant \cite{wang2016temporal} learn temporal evolution by using jointly the optical flow stream and the RGB stream for video classification. The recent video models \cite{yue2015beyond,donahue2015long,carreira2017quo,li2018videolstm} leverage long short-term memory (LSTM) to fuse frame-level CNN representations for modeling long-term temporal relationships. However, 2D CNN$+$LSTM \cite{carreira2017quo} empirically shows lower performance than two-stream architectures.
CNNs employing 3D convolution processing \cite{taylor2010convolutional,tran2015learning,carreira2017quo} represent a promising research direction for spatio-temporal representation learning, but the training of 3D CNNs has huge computational demands. Some research studies have devoted to simplifying 3D CNNs, such as P3D \cite{qiu2017learning}, TSM \cite{lin2019tsm}, S3D \cite{xie2018rethinking}, CSN \cite{tran2019video},  X3D \cite{feichtenhofer2020x3d}. Nevertheless, the inefficiency of message delivery caused by the deep stacking of local operations in 3D CNNs remains serious, and there is not much research on this problem, which is the main theme of this paper. \\

\noindent \textbf{Attention Mechanisms.} Attention mechanisms have been initally used for machine translation \cite{bahdanau2015neural}. Recent works \cite{Hu_2018_CVPR, wang2017residual, wang2018non, woo2018cbam} would embed task-specific attention mechanisms to CNNs to boost up performance and robustness in visual tasks. In computer vision, attention mechanisms can be decomposed into two components, channel attention - focusing on 'what' is meaningful, and spatial (or spatio-temporal) attention - focusing on 'where' is informative \cite{woo2018cbam}. For example, The Squeeze-and-Excitation (SE) module is a representative channel attention mechanism, which utilizes global average-pooled features to exploit the inter-channel relationships. Inspired by the classic non-local mean algorithm \cite{buades2005non} for image denoising, Wang, {\em et al.} \cite{wang2018non} introduced the self-attention concept  \cite{vaswani2017attention} from machine translation to large-scale visual classification tasks, and proposed non-local (NL) operation for video classification. The NL operation was initially designed to learn spatio-temporal attention. However, Cao {\em et al.} \cite{cao2019gcnet} observe that NL can only capture the global context of channels, aka channel attention. Moreover, they demonstrate that the intrinsic natures of the NL operation and SE module \cite{Hu_2018_CVPR} are the same while the implementation of the SE module is rather economical.

In this paper, we redesign the non-local operation and propose the region-based non-local operation which increases the effectiveness and efficiency in capturing the spatio-temporal attention. Yue {\em et al.} \cite{yue2018compact} also aimed to improve the NL operation, proposing a compact generalized version of the NL operation by integrating channel attention and spatio-temporal attention into a compact module. However, their work do not improve the effectiveness of NL operation. Instead of simplifying the NL, we focus on improving the effectiveness of NL for better capturing the spatio-temporal attention.

\section{Non-local Methods for Video Classification}
\label{sec3}

 \begin{figure*}
%   \centering
    \vspace{0pt}
    \vspace*{-0.5cm} 
  \begin{tabular}{c}
  \vspace{0pt}
%   \resizebox{0.9\textwidth}{!}{\input{nl.pdf_tex}}
    % \resizebox{1\textwidth}{!}{\input{nl.eps}}
    % \resizebox{1.\textwidth}{!}{\includesvg{nl.svg}}
    \resizebox{1\textwidth}{!}{\includegraphics{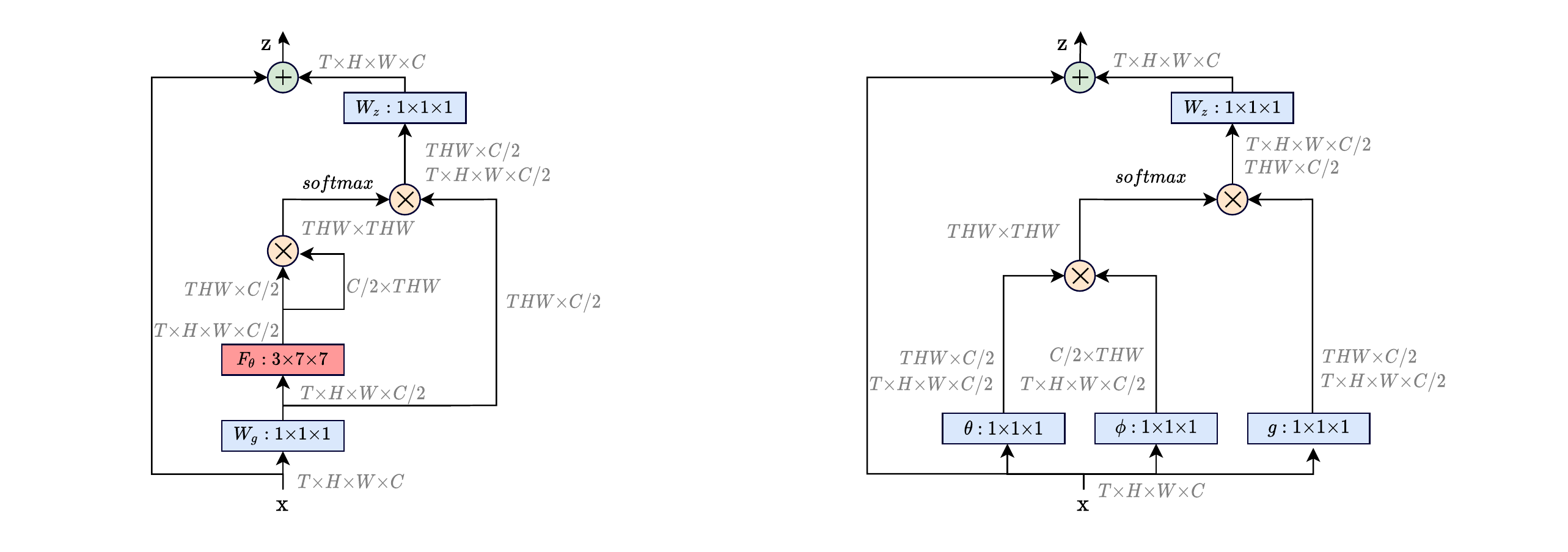}}
    % \resizebox{1.\textwidth}{!}{\includesvg{nl.svg}}
    
  \end{tabular}
  
  \begin{tabular}{l}
  \vspace{0pt}
   \hspace{3.cm} (a) RNL block  \hspace{6.5cm} (b) NL block \cite{wang2018non}
  \vspace*{-0.3cm}
  \end{tabular}
  \caption{Diagrams of implementing the NL and RNL operations in (b) and (a), respectively, indicating the shaping and the reshaping operations of a tensor together with the connections. $\otimes $ denotes matrix multiplication while $\oplus $ denotes element-wise addition. The blue boxes denote $1 \times 1 \times 1$ convolutions, and the red box $F_{\theta}$ denotes a $3 \times 7 \times 7 $ channel-wise convolution or an average/max pooling layer.}
  \vspace*{-0.3cm} 
\label{fig:nonlocal}
\end{figure*}

\subsection{Revisiting the Non-local (NL) Operation}
\label{sec_revisiting_nl}

Intuitively, the non-local operation \cite{wang2018non}, illustrated in Figure~\ref{fig:nonlocal} (b), strengthens the feature in a certain position via aggregating the information from other positions. The estimated value for a position, is computed as a weighted sum of the feature values of all other positions. Formally, we denote \(\mathbf{x}, \mathbf{y} \in \mathbb{R}^{THW \times C}\) as the input and output of an NL operation, flattened along the space-time directions, where $T$, $H$, $W$ and $C$ are temporal length (depth), height, width and the number of channels, respectively. Then, the NL operation can be described as:
\begin{equation} 
    \begin{aligned}
    \mathbf{y}_i\hspace{0.15cm} &= \frac{1}{ \mathcal{C}(\mathbf{x})}\sum_{\forall j}w_{i,j} \mathbf{W}_g \mathbf{x}_j,\\
    w_{i,j} &= f(\mathbf{x}_i, \mathbf{x}_j),
    \end{aligned}
    \label{eq_original_nl}
\end{equation}
where $\mathbf{x}_{i}, \mathbf{x}_{j} \in \mathbb{R}^{C}$ are the $i$-th and $j$-th element of $\mathbf{x}$, $i$ is the index of a reference position, and $j$ enumerates all possible positions. $\mathbf{W}_g$ is a learnable weight matrix that computes a representation of $\mathbf{x}_{j}$, and $\mathcal{C}(\mathbf{x})$ is the normalization factor. Meanwhile, $w_{i,j}$ is a weight, representing the relationship between positions $i$ and $j$, which is calculated by pairwise similarity function $f(\cdot,\cdot)$. Regarding the form for $f(\cdot, \cdot)$, Wang {\em et al.} \cite{wang2018non} propose four instantiations for the non-local operation, of which the embedded Gaussian form is described as $f(\mathbf{x}_i,\mathbf{x}_j) = e^{\theta(\mathbf{x}_i)^\mathsf{T} \phi(\mathbf{x}_j) }$, $\mathcal{C}(\mathbf{x})= \Sigma _{\forall j}f(\mathbf{x}_i,\mathbf{x}_j)$, where $\theta$ and $\phi$ represent linear transformations, implemented with $1\times1\times1$ convolutions.\\

\noindent \textbf{Attention Maps of the Non-local Operation.} In the NL operation, each output element $\mathbf{y}_i$ is a weighted average of the input features over all positions $\mathbf{x}_j$, and therefore each $\mathbf{y}_i$ has a corresponding attention weight map calculated by $f(\cdot,\cdot)$, highlighting the areas related to position $i$. In Figure~\ref{fig:example_nl} (b), we randomly pick one video from Kinetics-400 and visualize the attention maps of NL at two different reference positions, one of which is located in the background area while the other is located in the region of the moving object. In the original NL operation, its attention maps with different reference positions are almost the same, which indicates that this fails to capture the positional relations. The NL operation realistically learns channel-wise attention rather than spatio-temporal attention. 

We redesign the non-local operation as a spatio-temporal attention mechanism, namely the region-based non-local operation (RNL). Figure~\ref{fig:example_nl} (a) shows that our RNL operation only highlights the regions related to the reference position, which indicates that the proposed RNL operation can effectively learn spatio-temporal attention.

% \vspace*{-0.1cm}
\subsection{Region-based non-local (RNL) Operation}
% \vspace*{-0.1cm}
\label{region-based non-local operation}

The initial idea for the RNL operation is that the relation between two positions in a video representation should not rely on just their own features but also on those features from their neighborhoods. Therefore, for each position $i$ of input sample $\mathbf{x}$, we define a cuboid region $\mathcal{N}_i$ of fixed size centered at position $i$. The calculation of the relationship $w_{i,j}$ between positions $i$ and $j$ is redefined as:
% \vspace*{-0.2cm}
\begin{equation}
\begin{aligned}
    w_{i,j} &= f(\theta(\mathcal{N}_i), \theta(\mathcal{N}_j)),
    \label{eq_fn_f}
\end{aligned}
% \vspace*{-0.2cm}
\end{equation}
where, $\theta (\cdot)$ denotes an information aggregation function that separately summarizes the features in a region for each channel. Function $\theta (\cdot)$ is given by
\vspace*{-0.15cm}
\begin{equation}
\begin{aligned}
    \theta(\mathcal{N}_i) = \sum_{k \in \mathcal{N}_i}\mathbf{u}_k \odot \mathbf{x}_k,
    \label{eq_fn_theta}
\end{aligned}
\vspace*{-0.2cm}
\end{equation}
where $\odot$ denotes element-wise multiplication and $\mathbf{u}_k$ denotes a vector shared by all cuboid regions $\mathcal{N}_i$. As there is no channel interaction in $\theta (\cdot)$, it can be implemented as channel-wise \footnote{Also referred to as “depth-wise". We use the term “channel-wise" to avoid confusions with the network depth.} convolutions \cite{sandler2018mobilenetv2}, or as average/max pooling. By replacing the expression of $w_{i,j}$ from equation  (\ref{eq_original_nl}) with the expression from (\ref{eq_fn_f}), the RNL operation can be written as:
\begin{equation} 
    \mathbf{y}_{i}=\frac{1}{\mathcal{C}(\mathbf{x})}\sum_{\forall j}f(\theta(\mathcal{N}_i), \theta(\mathcal{N}_j))\mathbf{x}_{j}.
    \label{eq_inl_y}
\end{equation}

From equation (\ref{eq_inl_y}), we can see that by employing the RNL operation, the new feature of each position is a weighted sum of the old features from all positions, where the weights are calculated by the similarity function $f(\cdot, \cdot)$ according to the similarity between the target region, and all the other regions. The proposed RNL operation enhances the calculation of positional relations by fully utilizing the information from the neighboring regions, which increases the robustness to noise or unrelated features, Hence, the RNL operation can learn more meaningful representations in comparison with NL.

For the form of function $f(\cdot,\cdot)$, in addition to adopting the Gaussian version and the Dot product version as in \cite{wang2018non}, we also propose a new form, called the Cosine version. Specifically, the \textbf{Gaussian} form of $f(\cdot,\cdot)$ is given by

\begin{equation}
f(\theta(\mathcal{N}_i), \theta(\mathcal{N}_j)) = e^{\theta(\mathcal{N}_i)^ \mathsf{T} \theta(\mathcal{N}_j)}.
\label{eq_f_gaussian}
\end{equation}

The \textbf{Dot product} form of $f(\cdot,\cdot)$ measures the relation between two regions by using the dot-product similarity:

\begin{equation} 
   f(\theta(\mathcal{N}_i), \theta(\mathcal{N}_j)) = \theta(\mathcal{N}_i)^ \mathsf{T} \theta(\mathcal{N}_j).
    \label{eq_dot_product}
\end{equation}

However, the dot-product similarity takes into account both the vector angle and the magnitude, as  $\theta(\mathcal{N}_i)^\mathsf{T} \theta(\mathcal{N}_j)= \|\theta(\mathcal{N}_i)   \|  \|\theta(\mathcal{N}_j)  \| \cos \psi_{i,j}$, where $\psi_{i,j}$ is the angle between vectors $\theta(\mathcal{N}_i)$ and $\theta(\mathcal{N}_j)$. It is preferable to replace dot-product similarity with the cosine similarity, ignoring the vector magnitude and resulting in a value within the range $[-1,1]$. The \textbf{Cosine} form of $f(\cdot,\cdot)$ is expressed as:
\begin{equation}
\begin{array}{l}
   f(\theta(\mathcal{N}_i), \theta(\mathcal{N}_j)) = \mathrm{ReLU}(\frac{\theta(\mathcal{N}_i)^\mathsf{T} \theta(\mathcal{N}_j)}{ \|\theta(\mathcal{N}_i)  \|  \|\theta(\mathcal{N}_j) \|})\\
  \hspace{2.47cm}  = \mathrm{ReLU}(\cos \psi_{i,j}).
\end{array}
\label{eq_cos}
\end{equation}

When $f(\theta(\mathcal{N}_i), \theta(\mathcal{N}_j))<0$, it indicates that the features in positions $i$ and $j$ are not related. As the new feature in a cetrain position should only be determined by those related features, we use the ReLU function to restrict the output of $f(\cdot, \cdot)$ to be non-negative. The normalization factor is set as $\mathcal{C(\mathbf{x})} = \sum_{\forall j}f(\theta(\mathcal{N}_i), \theta(\mathcal{N}_j))$ for the Gaussian version from (\ref{eq_f_gaussian}), and set as $\mathcal{C(\mathbf{x})} = THW$ for the Dot-product and Cosine versions from equations (\ref{eq_dot_product}) and (\ref{eq_cos}), respectively.

\subsection{Region-based non-local Block}
\label{region-based non-local Block}

In order to embed the RNL operation into the off-the-shelf CNNs without influencing the results provided by the pre-trained kernels, we embed the RNL operation into a residual style block \cite{he2016deep}, named the RNL block. The Gaussian RNL block, defined by (\ref{eq_f_gaussian}), is written as a matrix form as:
\begin{align}
\vspace*{-0.5cm}
\mathbf{z}&=\mathbf{y}\mathbf{W}_z + \mathbf{x}, 
\label{eq_nl_block_0}\\
\mathbf{y}&={softmax}(F_\theta(\mathbf{x}\mathbf{W}_g) (F_\theta(\mathbf{x}\mathbf{W}_g))^{\mathsf{T}}) \mathbf{x}\mathbf{W}_g,
\label{eq_nl_block_1}
\end{align}
where $\mathbf{z}$ is the output that represents the feature after recalibration, $\mathbf{W}_{z} \in \mathbb{R}^{\frac{C}{2} \times C}$ and $\mathbf{W}_{g} \in \mathbb{R}^{C \times \frac{C}{2}}$ are learnable weight matrices, which are implemented as $1\times1\times1$ convolutions, and '$+ \mathbf{x}$'  denotes a residual term. 
$F_\theta$ denotes the operation that corresponds to the matrix form of function $\theta(\cdot)$ from equation (\ref{eq_fn_theta}). We present the architectures of the Gaussian RNL block and the Gaussian embedding version of the original NL block in Figure~\ref{fig:nonlocal}. We can observe that the original NL block illustrated in Figure \ref{fig:nonlocal} (b) uses four $1 \times 1 \times 1$ convolutions, while the proposed RNL block shown in Figure \ref{fig:nonlocal} (a) uses two $1 \times 1 \times 1$ convolutions and one channel-wise convolution, which reduces the computational complexity significantly. 
\begin{figure}
%   \centering
  \begin{tabular}{c}
    % \resizebox{0.4\textwidth}{!}{\includesvg{depthwise.svg}}
    % \resizebox{0.4\textwidth}{!}{\input{depthwise.pdf_tex}}
    \resizebox{0.44\textwidth}{!}{\includegraphics{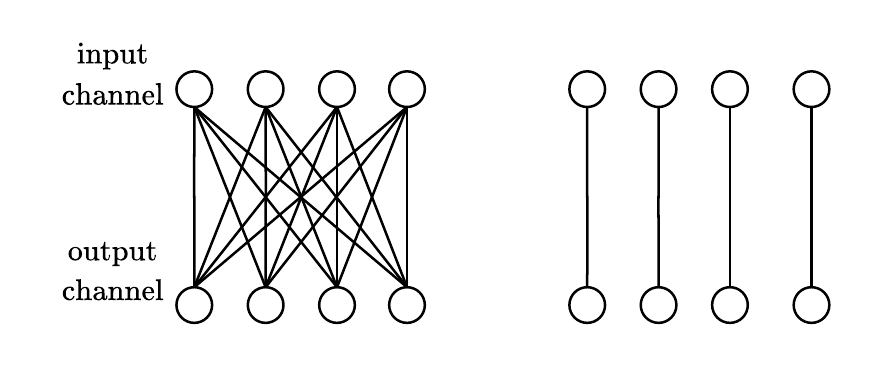}}
  \end{tabular}
  \begin{tabular}{l}
  (a) conventional convolution \hspace{0.1cm} (b) channel-wise convolution
  \end{tabular}
  \caption{Illustrations of the conventional convolution (a) and the channel-wise convolution (b). The total number of connections of the channel-wise convolution \cite{sandler2018mobilenetv2} is reduced to $\frac{1}{C}$ of that of the conventional convolution.}
\label{fig:depthwise}
\end{figure}

Next, we explain two main implementations of the region information aggregation function $F_{\theta}$ in RNL operation. 

\textbf{1) Channel-wise Convolutions.} It is worthwhile to note that, in principle, the candidates for implementing $F_{\theta}$ should not fuse together information across channels. Otherwise, the new feature embedding might fail to represent its original information, which is why we cannot adopt conventional convolutions. In contrast, channel-wise convolution \cite{sandler2018mobilenetv2}, exemplified in Figure \ref{fig:depthwise}, is a perfect candidate for the implementation of $F_{\theta}$, as there is no interaction between the channels. An additional benefit that the channel-wise convolution brings is that it reduces the computation and the parameters by a factor of $C$, compared with the conventional convolution. The kernel size of the channel-wise convolution has a significant impact on performance, as it corresponds to how large a region $\mathcal{N}_i$ is considered for information aggregation. We will explore the effectiveness of various kernel sizes, in Section \ref{sec_exploration}.

\textbf{2) Average/Max Pooling.} The other implementation options for $F_\theta$ are the average pooling and max pooling, which have been widely adopted for information aggregation. Although it shows a relatively weaker capability than the implementation of channel-wise convolution, average/max pooling adds no extra parameters to the models.

\begin{table}
\centering
\caption{The architecture of the RNL network. The kernel size  and the output size are shown in the second and third columns, respectively. The RNL blocks are inserted after the residual blocks shown in brackets, where the temporal shift modules \cite{lin2019tsm} are embedded into the convolutional layers.}
\label{tabel:archi}
\setlength{\tabcolsep}{11pt}
\begin{tabular}{c|l|c}
% \hline
Layer & Operation & Output size\\
\thickhline
conv1 & $1\times7\times7$, 64, stride 1,2,2& $8\times112\times112$   \\
\hline
pool1 & $1\times3\times3$, 64, stride 1,2,2& $8\times56\times56$  \\
\hline
res2 & $\begin{bmatrix}
   1 \times 1 \times 1, 64  \\
   1 \times 3 \times 3, 64 \\
   1 \times 1 \times 1, 256 
\end{bmatrix} \hspace{0.85cm} \times 3$& $8\times56\times56$  \\
\hline
res3 & $\begin{bmatrix}
   1 \times 1 \times 1, 128  \\
   1 \times 3 \times 3, 128 \\
   1 \times 1 \times 1, 512 
\end{bmatrix}_ \mathrm{RNL} \hspace{0.25cm} \times 4$ & $8\times28\times28$  \\
\hline
res4 & $\begin{bmatrix}
   1 \times 1 \times 1, 256  \\
   1 \times 3 \times 3, 256 \\
   1 \times 1 \times 1, 1024 
\end{bmatrix}_ \mathrm{RNL} \hspace{0.05cm} \times 6$ & $8\times14\times14$  \\
\hline
res5 & $\begin{bmatrix}
   1 \times 1 \times 1, 512  \\
   1 \times 3 \times 3, 512 \\
   1 \times 1 \times 1, 2048 
\end{bmatrix} \hspace{0.65cm} \times 3$ & $8\times7\times7$  \\
% \hline
\end{tabular}
\end{table}

\vspace*{-0.1cm}
 \subsection{Attention Chain}
 \label{sec_att_chain}

When the proposed RNL block can learn the long-range dependencies for each position in the spatio-temporal dimension, the squeeze-excitation (SE) block \cite{Hu_2018_CVPR} can learn the long-range dependencies in the channel dimension. In order to capture both spatio-temporal attention and channel-wise attention in a single module, we embed the SE block \cite{Hu_2018_CVPR} together with the RNL block to form an attention chain module (SE+RNL). Firstly, we modify the SE block \cite{Hu_2018_CVPR}, where the squeeze operation $F_{sq}$ is expressed as:
\begin{equation} 
   \mathbf{s}' = F_{sq}(\mathbf{x})= \frac{1}{THW}\sum_{i=1}^{THW}\mathbf{x}_i,
    \label{eq_squeeze}
\end{equation}
and the excitation operation $F_{ex}$ is expressed as:

\begin{equation} 
   \mathbf{s} = F_{ex}(\mathbf{s}')  = \mathbf{W}_2\mathrm{ReLU}(\mathrm{BN}(\mathbf{W}_1\mathbf{s}')),
    \label{eq_excitation}
\end{equation}
where $\mathbf{W}_1 \in \mathbb{R}^{\frac{C}{2} \times C}$ and $\mathbf{W}_2 \in \mathbb{R}^{C \times \frac{C}{2}}$ are learnable weights, which can be implemented with fully-connected (FC) layers. In the excitation operation $F_{ex}$, we add a batch normalization ($\mathrm{BN}$) layer \cite{szegedy2015bn} right after the FC layer $\mathbf{W}_1$ to reduce the internal covariate shift. Subsequently, we reshape $\mathbf{s} \in \mathbb{R}^{C}$ into $\mathbb{R}^{1\times C}$. The output of the SE block is given by:
\begin{equation} 
   \mathbf{v} = \mathbf{x} \oplus \mathbf{s},
    \label{eq_se_output}
\end{equation}
where $\oplus$ refers to the element-wise addition broadcasting in unmatched dimensions (replicate $\mathbf{x}$ to match the dimension of $\mathbf{s}$). After that, we place the RNL block after the SE block to form an attention chain.

% \vspace*{-0.1cm}
\subsection{The Network Architecture}

The RNL block is designed to be compatible with most existing CNNs. It can be plugged into a CNN at any processing stage, resulting in an RNL network. For the implementation, we use ResNet-50 \cite{he2016deep} with the temporal shift modules (TSM) \cite{lin2019tsm} as the backbone network to build our model (RNL TSM), and its structures is provided in Table~\ref{tabel:archi}. The TSM is a lightweight module enabling 2D CNNs to achieve temporal modeling by shifting part of the channels along the temporal dimension, which facilitates the information exchange among neighboring frames. In this architecture, we keep the temporal size constant, which means all the layers in the network only reduce the spatial size of the input features. The backbone network is also the baseline for our experiments.

\section{Experiments}

 \begin{figure*}
    \centering
    \resizebox{0.95\textwidth}{!}{\includegraphics{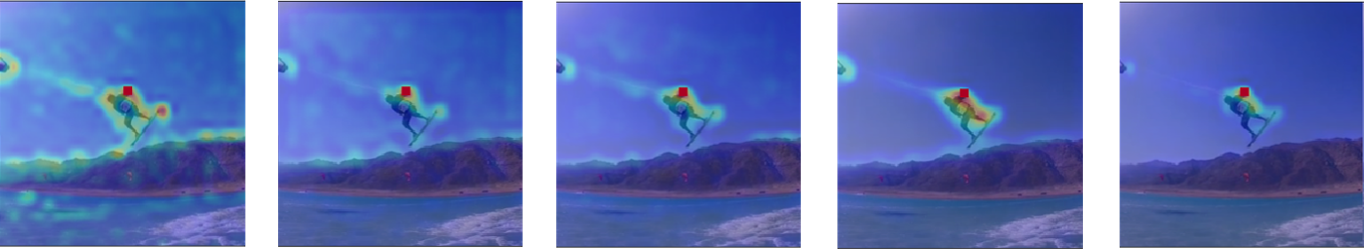}}\\
    \small$1 \times 1$ \hspace{2.5cm} $3 \times 3$ \hspace{2.5cm}  $5 \times 5$ \hspace{2.5cm} $7 \times 7$ \hspace{2.5cm} $9 \times 9$
    \caption{Visualization the attention maps of the RNL block when considering different kernel sizes in the res3 stage by giving the reference position (red point). When the reference point is located at the moving object, the RNL operation with proper kernel size should just highlight the related moving regions. 
    % \vspace*{-0.4cm}
    }
    \label{fig:ks_examples}
\end{figure*}

\begin{table*}
    \makeatletter\def\@captype{table}\makeatother\caption{Exploration of the effectiveness and efficiency of various RNL modules on Kinetics-400. For the models in (a) and (c), we insert one Gaussian RNL block into the res3 stage of ResNet-50.}
    % \vspace*{-0.2cm}
    \label{table:exploration}
    % \centering
    \begin{minipage}[t]{0.001\textwidth}
    \quad
    \end{minipage}
    \begin{minipage}[t]{0.3\textwidth}
    \vspace{0pt}
    % \vspace*{-1.539cm}
    \setlength{\tabcolsep}{5pt}
    \begin{tabular}{cc|cc}
    Kernel size & Top-1 (\%) & Kernel size & Top-1 (\%)\\
    \thickhline
    $1 \times1 \times 1$ & 73.28 & $3 \times 3 \times 3$ & 73.53 \\
    $3 \times 1 \times 1$ & 73.41 & $3 \times 5 \times 5$ & 73.27 \\
    $7 \times 1 \times 1$ & 73.12 & $\mathbf{3 \times 7 \times 7}$ & \textbf{73.66} \\
    $1 \times 3 \times 3$ & 73.32 & $3 \times 9 \times 9$ & 73.51\\
    $1 \times 7 \times 7$ & 73.43 & $7 \times 7 \times 7$ & 73.11 \\
    $1 \times 9 \times 9$ & 73.32 & $7 \times 9 \times 9$ & 73.30 \\
    \end{tabular}\\
    % \vspace*{-0.09cm}
    (a) RNL blocks with different kernel sizes of $F_\theta$.
    \end{minipage}
    \begin{minipage}[t]{0.05\textwidth}
    \quad
    \end{minipage}
    % \vspace*{+0.3cm}
    \begin{minipage}[t]{0.25\textwidth}
    % \vspace*{+0.02cm}
    \vspace{0pt}
    \begin{tabular}{ccc}
    \# RNL & Method($f(\cdot,\cdot)$) & Top-1 (\%)  \\
    \thickhline
    &Dot-product & 73.22\\ 
    1&Gaussian & \textbf{73.66}\\ 
    &Cosine & 73.46\\ 
    \hline
    &dot-product & 74.16\\ 
    5&Gaussian & \textbf{74.68}\\ 
    &Cosine & 74.40
    % \vspace*{+0.3cm}
    \end{tabular}\\
    (b) Instantiations of the RNL with different form of $f(\cdot, \cdot)$.\\
    \end{minipage}
    \begin{minipage}[t]{0.03\textwidth}
    \quad
    \end{minipage}
    \begin{minipage}[t]{0.3\textwidth}
    \vspace{0pt}
    \setlength{\tabcolsep}{3pt}
    \begin{tabular}{lccc}
    % \hline
    Method ($F_\theta$) & Top-1 (\%) & GFLOPs & Params \\
    \thickhline
    channel-wise conv & 73.66 & 1.65 & 2.67M\\
    average pooling & 73.22 & 1.65 & 0.26M \\
    max pooling & 73.47 & 1.65 & 0.26M \\
    \\
    \\
    \\
    % \vspace*{+1.cm}
    \end{tabular}\\
    (c) Instantiations of RNL with different implementations of $F_\theta$. 
    \end{minipage}
    \vspace*{-0.5cm}
\end{table*}

\begin{figure*}
  \centering
  \begin{tabular}{c}
    \includegraphics[width=.97\textwidth]{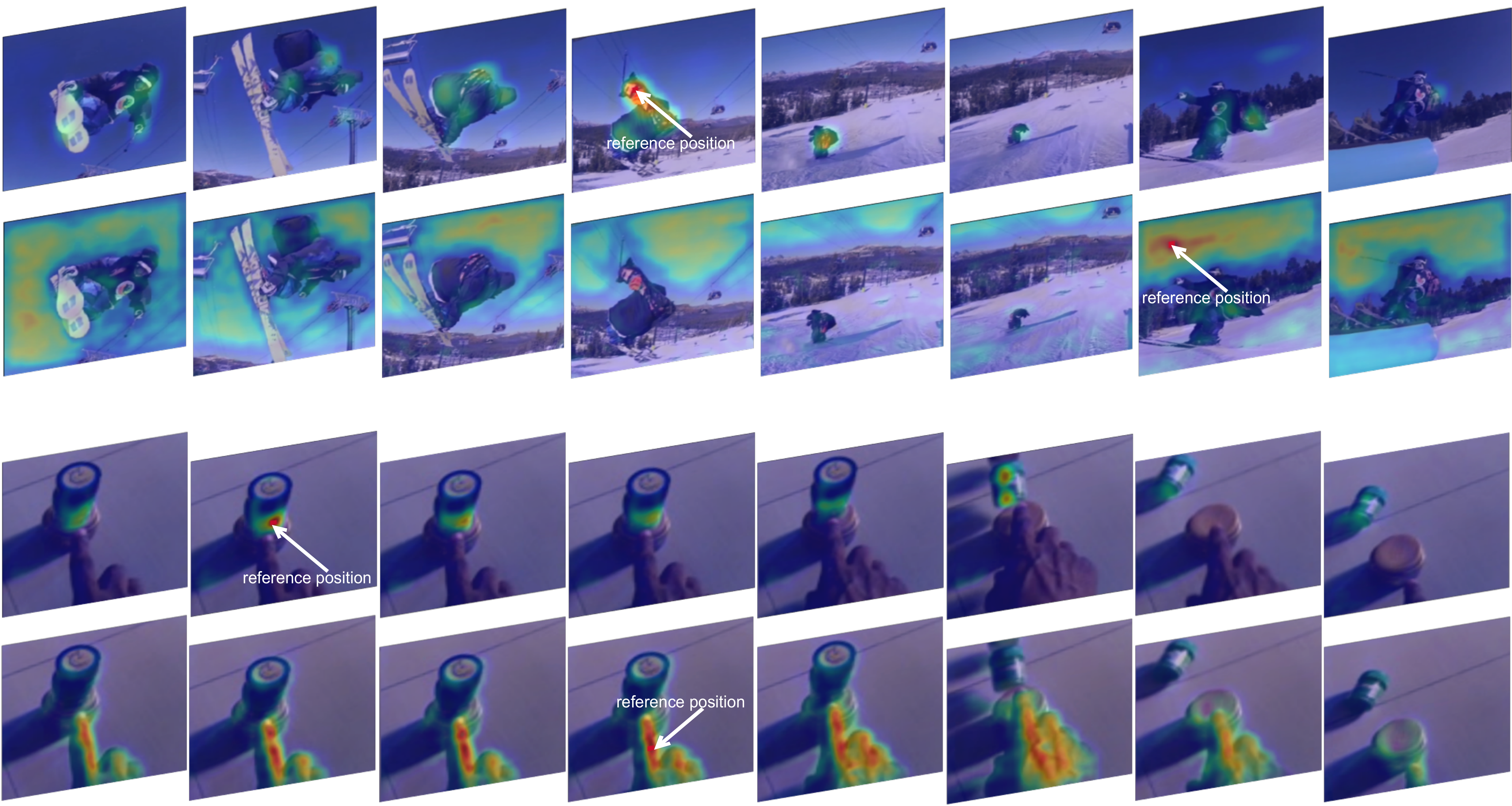}
  \end{tabular}
 
  \caption{Visualization of attention maps of the RNL in the res3 stage, with different reference positions on frames from Kinetics (1st row) and Something-Something (2nd row). Given a video clip,  the RNL operation only highlights those regions related to the reference position.}
\label{fig:example_nl_some}
\end{figure*}

We perform video classification experiments on two standard video benchmarks, Kinetics-400 \cite{carreira2017quo} and Something-Something V1 \cite{goyal2017something}. Kinetics-400 is a large-scale video classification benchmark that consists of $\sim$300K video clips, classified into 400 categories. Something-Something V1 consists of $\sim$108K videos from 174 categories. We report Top-1, Top-5 accuracy on the validation sets and the computational cost (in GFLOPs) of a single, spatially center-cropped clip to comprehensively evaluate the effectiveness and efficiency. Figure~\ref{fig:example_nl} and Figure~\ref{fig:example_nl_some} visualize some examples of the attention maps of RNL operation, which shows RNL operation can correctly learn the relations between positions. 

\noindent \textbf{Training and Inference.} Our models are pretrained on ImageNet \cite{imagenet_cvpr09}. For the training, we follow the setting from \cite{wang2018non} and use a spatial size of 224 $\times$ 224, which is randomly cropped from a resized video frame. The temporal size is set as 8 frames unless otherwise specified. In order to prevent overfitting, we add a dropout layer after the global pooling layer.
We optimize our models using the Stochastic Gradient Descent, and train the models for 50 epochs with a cosine decay learning rate schedule. The batch size is set at 64 across multiple GPUs. For Kinetics, the initial learning rate, weight decay and dropout rate are set to 0.01, 1e-4 and 0.5 respectively; for Something-Something, these hyper-parameters are set to 0.02, 8e-4, and 0.8 respectively. In the inference, we follow the common setting in \cite{wang2018non,lin2019tsm}. Unless stated otherwise, we uniformly sample 10/2 clips for Kinetics-400/Something-Something V1, and perform spatially fully convolutional inference (three crops of size 256 $\times$ 256 to cover the spatial dimensions) for all clips, and the video-level prediction is obtained by averaging all the clip prediction scores of a video.

\subsection{Ablation Studies}

\label{sec_exploration}
 We explore the most efficient and effective form of RNL operation on Kinetics-400. By default, the function $f(\cdot, \cdot)$ of RNL operation is implemented by using the equation (\ref{eq_f_gaussian}), and $F_\theta$ is implemented by a channel-wise convolution with a kernel size of $3 \times 7 \times 7$, unless otherwise specified. Following the results from \cite{wang2018non}, we add RNL blocks to the res3 and res4 stages in the architecture shown in Table~\ref{tabel:archi}.
 Our exploration is organized in three parts. First, we search for the effective kernel size of $F_{\theta}$ in RNL blocks. Next, we evaluate the performance of various instantiations of RNL and find out the efficient and effective one. 
Finally, we combine the selected version of RNL with an SE block to form an attention chain module.

\noindent \textbf{Kernel Size.} The kernel size of $F_\theta$ (determining the size of region $\mathcal{N}_i$) in the RNL block has a significant impact on the performance as it affects what the RNL operation would learn. Large kernels are supposed to be robust to noise, while small kernels would consider the details and fine structures from video sequences. By considering that the features learned by the kernel from the temporal and spatial dimensions are different, we separately explore the temporal and spatial sizes of the kernel by fixing one while varying the other. The results are shown in Table~\ref{table:exploration} (a). We observe that in the temporal dimension, the size of 3 surpasses other options regardless of the spatial size of the kernel, while in the spatial dimension, the size of 7 is the best option. Therefore, we expect the kernel of $3 \times 7 \times 7$ is the best option in space and time, and it has been verified through our grid search. Concurrently, we evaluate the influence of the kernel size of $F_\theta$ to the model performance by visualizing the attention maps of the RNL operation, shown in Figure~\ref{fig:ks_examples}, where the RNL operation considers the highlighted areas to have strong relations with the reference position, indicated by a red point. Figure~\ref{fig:ks_examples} shows that a kernel of a small size spatially, such as $1 \times 1$, tends to incorrectly interpret the relations between some background areas and the foreground areas. In contrast, a kernel with larger spatial size can learn more precise relations between such positions. For example, the kernel of size $7\times7$ precisely highlights the moving object in in Figure~\ref{fig:ks_examples} when the reference position is located at the moving object. However, too large kernels could also lead to performance degradation. For example, the kernel of size $3 \times 9 \times 9$ has a lower accuracy than the kernel of size $3 \times 7 \times 7$ (73.51\% vs. 73.66\%), and the kernel of $7 \times 7 \times 7$ shows a lower performance than the kernel of size $3 \times 7 \times 7$ (73.11\% vs. 73.66\%). The kernel of size $1\times1\times1$ has a lower accuracy than the others except for $7\times1\times1$ and $7 \times 7 \times 7$, which verifies our assumption that the relation between two positions should not rely on just their own features but also on features from their neighborhoods.

\noindent \textbf{Instantiations.} There are various solutions for $f(\cdot, \cdot)$ from equation (\ref{eq_inl_y}) and for $F_\theta$ from equation (\ref{eq_nl_block_1}), as discussed in Section \ref{region-based non-local operation} and Section \ref{region-based non-local Block}, respectively. In the following, we conduct ablation studies on the instantiations by fixing a specific choice for either $f(\cdot, \cdot)$ or $F_\theta$ while changing the other. The operation $F_\theta$ can be implemented as a channel-wise convolution or as the average/max pooling, the stride of which is set as 1, and the padding of which is half of the kernel size. From the results shown in Table~\ref{table:exploration} (c), we can see that the channel-wise convolution implementation achieves a higher accuracy with +0.44\% and  +0.19\% than the average and max pooling, respectively. However, the implementation of average/max pooling is more efficient and adds fewer parameters (-2.4M) to the model compared to the channel-wise convolution. We instantiate three versions of the RNL operation, such as Gaussian, Dot-product and Cosine, provided in equations (\ref{eq_f_gaussian}), (\ref{eq_dot_product}) and (\ref{eq_cos}) respectively. The results are shown in Table~\ref{table:exploration} (b). By adding a single RNL block into the backbone network, the result of the Gaussian RNL outperforms the Dot-product and Cosine versions. Moreover, the performance of all installations of the RNL operation can be further improved by stacking more RNL blocks. The model with 5 Gaussian RNL blocks (3 in the res4 stage and 2 in the res3 stage) gains an additional 1.02\% accuracy increase in comparison with adding a single RNL block.

\subsection{Evaluation}

 In order to evaluate the efficiency and effectiveness of our method in comparison with other attention mechanisms, we reimplement the original NL network \cite{wang2018non}, GCNet \cite{cao2019gcnet} (a simplified NL network), SE network \cite{Hu_2018_CVPR} and CBAM network \cite{woo2018cbam}.  Table~\ref{table:comparisons_attention_mechanisms} presents the results on Kinetics and Something-Something. We can see that the proposed RNL block achieves higher performance than other attention mechanisms. Notably, the network with 5 RNL blocks outperforms the network with 5 NL blocks with +0.27\% on Kinetics and +1\% on Something-Something, while the computational complexity required in FLOPs of the RNL network is 8.23G less than that of the NL network. Furthermore, by adding 5 blocks of the attention chain (SE + RNL), as described in Section~\ref{sec_att_chain}, to the backbone network, the performance is further improved (74.97\% on Kinetics and 49.47\% on Something-Something). In the visualization examples of the RNL and NL blocks, shown in Figure~\ref{fig:example_nl}, we observe that the attention maps of the RNL block would only highlight those regions related to the reference positions. However, the attention maps of the original NL block always highlight the same regions for different reference positions. The observation demonstrates that the RNL block can capture the spatio-temporal attention while the NL block only captures the channel attention. 
 
  \begin{table}
\caption{Comparisons between various visual attention mechanisms on Kinetics-400 and Something-Something V1.}
\label{table:comparisons_attention_mechanisms}
\centering
\setlength{\tabcolsep}{4pt}
\begin{tabular}{l|lccc}
Dataset & Model & Top-1 (\%)  & FLOPs (G)  & \# Param (M) \\
\thickhline
&baseline & 72.80 & 32.89 & 24.33\\
&+ 5 SE & 73.70 & 32.89 &24.79\\
Kinetics-&+ 5 CBAM & 73.99 & 32.90 & 24.80 \\
400&+ 5 GC & 73.76 & 32.90 & 24.79\\
&+ 5 NL  & 74.41 & 49.38 & 31.69\\
&\textbf{+ 5 RNL} & \textbf{74.68} & \textbf{41.15}& \textbf{35.48}\\
&\textbf{+ 5 $[$SE+RNL$]$} & \textbf{74.97} & \textbf{41.16}& \textbf{35.95}\\
\hline
\hline
Something-& baseline & 46.63 & 32.89 & 24.33\\
Something & + 5 NL & 48.25 & 49.38 & 31.69 \\
V1& \textbf{+ 5 RNL} & \textbf{49.24} & \textbf{41.15}& \textbf{35.48}\\
&\textbf{+ 5 $[$SE+RNL$]$} & \textbf{49.47} & \textbf{41.16}& \textbf{35.95}\\
\end{tabular}
% \vspace*{-0.4cm}
\end{table}

\begin{table}
\caption{Results on Kinetics-400.}
\label{table:sota}
\centering
\setlength{\tabcolsep}{7.8pt}
\begin{tabular}{llccc}
 Model & Backbone & Training & Top-1 & Top-5 \\
      &           & Frames    &  &\\
\thickhline
I3D RGB \cite{carreira2017quo} & Inception &64 & 72.1 & 90.3\\
% I3D Flow \cite{carreira2017quo} & Inception & 64 & 65.3 & 86.2\\
% I3D RGB+Flow \cite{carreira2017quo} & Inception & 64 & 75.7 & 92.0\\
S3D-G RGB \cite{xie2018rethinking} & Inception & 64 & 74.7 & 93.4\\
TSM \cite{lin2019tsm} & ResNet-50 & 8 & 74.1& 91.2 \\
 TSM \cite{lin2019tsm} & ResNet-50 & 16 & 74.7& - \\
% S3D-G Flow \cite{xie2018rethinking} & Inception & 64 & 68.0 & 87.6\\
% S3D-G RGB+Flow \cite{xie2018rethinking} & Inception & 64 & 77.2 & 93.0\\
NL I3D \cite{wang2018non} & ResNet-50 & 32 & 74.9 & 91.6 \\
Slow \cite{feichtenhofer2019slowfast} & ResNet-50 & 8 & 74.9 & 91.5\\
SlowFast \cite{feichtenhofer2019slowfast} & ResNet-50 & 4+32 & 75.6 & 92.1\\
\textbf{RNL TSM (ours)} & \textbf{ResNet-50} & \textbf{8} & \textbf{75.6} & \textbf{92.3} \\
\textbf{RNL TSM (ours)} & \textbf{ResNet-50} & \textbf{16} & \textbf{77.2} & \textbf{93.1} \\
\textbf{RNL TSM$_{En}$ (ours)} & \textbf{ResNet-50} & \textbf{8+16} & \textbf{77.4} & \textbf{93.2} \\
NL I3D \cite{wang2018non} & ResNet-50 & 128 & 76.5& 92.6 \\
NL I3D \cite{wang2018non} & ResNet-101 & 128 & 77.7& 93.3 \\
SlowFast \cite{feichtenhofer2019slowfast} & ResNet-101 & 16+64 & 78.9 & 93.5\\
LGD-3D RGB \cite{qiu2019learning} & ResNet-101 & 128 & 79.4 & 94.4 \\
\end{tabular}
\end{table}

\subsection{Comparisons with the State-of-the-Art}
We compare the proposed method with the state-of-the-art methods on Kinetics-400 and Something-Something V1. In order to achieve the best performance on Kinetics-400, we increase the number of training epochs from 50 to 100. The performance comparisons are summarized in Tables~\ref{table:sota} and \ref{table:sota_some}, where RNL TSM refers to the model with 5 attention chain blocks. Note that using the same approach, the models with deeper backbone networks or longer clips as training inputs would consistently result in better performance in comparison with shallower backbone networks. on Kinetics, we use a shallower network, such as ResNet-50, as the backbone, and the length of our input video clips is at least 8 times shorter than other methods, yet our results are highly competitive with those of the other approaches. 

\begin{table}
\caption{Results on Something-Something V1. }
\label{table:sota_some}
\centering
% \small
\setlength{\tabcolsep}{3.9pt}
\begin{tabular}{lllcc}
Model & Backbone & Frame$\times$Clip$\times$Crop & Top-1 & Top-5 \\
\thickhline
 I3D \cite{wang2018videos} & ResNet-50 & 192=32$\times$2$\times$3& 41.6& 72.2\\
 NL I3D \cite{wang2018videos} & ResNet-50 & 192=32$\times$2$\times$3 & 44.4 & 76.0\\
NL I3D + GCN \cite{wang2018videos} & ResNet-50 & 192=32$\times$2$\times$3 & 46.1 & 76.8\\
TSM \cite{lin2019tsm} & ResNet-50 & 8=8$\times$1$\times$1 & 45.6 & 74.2\\
TSM \cite{lin2019tsm} & ResNet-50 & 16=16$\times$1$\times$1 & 47.2 & 77.1\\
TSM$_{En}$ \cite{lin2019tsm} & ResNet-50 & 24=(8+16)$\times$1$\times$1 & 49.7 & 78.5\\
\textbf{RNL TSM (ours)} & \textbf{ResNet-50} & \textbf{8=8$\times$1$\times$1} & \textbf{47.3} & \textbf{-} \\
\textbf{RNL TSM (ours)} & \textbf{ResNet-50} & \textbf{16=16$\times$1$\times$1} & \textbf{49.4} & \textbf{-} \\
\textbf{RNL TSM$_{En}$ (ours)} & \textbf{ResNet-50} & \textbf{24=(8+16)$\times$1$\times$1} & \textbf{51.3} & \textbf{80.6} \\
\hline
SmallBig \cite{li2020smallbignet} & ResNet-50 & 48=8$\times$2$\times$3 & 48.3 & 78.1 \\
SmallBig \cite{li2020smallbignet}& ResNet-50 & 96=16$\times$2$\times$3 & 50.0 & 79.8 \\
SmallBig$_{En}$ \cite{li2020smallbignet} & ResNet-50 & 144=(8+16)$\times$2$\times$3 & 51.4 & 80.7 \\
\textbf{RNL TSM (ours)} & \textbf{ResNet-50} & \textbf{48=8$\times$2$\times$3} & \textbf{49.5} & \textbf{78.4} \\
\textbf{RNL TSM (ours)} & \textbf{ResNet-50} & \textbf{96=16$\times$2$\times$3} & \textbf{51.0} & \textbf{80.3} \\
\textbf{RNL TSM$_{En}$ (ours)} & \textbf{ResNet-50} & \textbf{144=(8+16)$\times$2$\times$3} & \textbf{52.7} & \textbf{81.5} \\
\hdashline
\textbf{RNL TSM (ours)} & \textbf{ResNet-101} & \textbf{48=8$\times$2$\times$3} & \textbf{50.8} & \textbf{79.8} \\
\textbf{RNL TSM$_{En}$ (ours)} & \textbf{R101 + R50} & \textbf{144=(8+16)$\times$2$\times$3} & \textbf{54.1} & \textbf{82.2} \\
\end{tabular}
% \vspace*{-0.27cm}
\end{table}

On Something-Something V1, when using ResNet-50 as the backbone, the ensemble version of our model, the RNL TSM$_{En}$, using \{8, 16\} frames as inputs, achieves a higher accuracy than other approaches, w.r.t., single-clip \& center-crop (Top-1: 51.3\%) and multi-clip \& multi-crop (Top-1: 52.7\%). When adopting ResNet-101 as the backbone, we gain extra performance boost (Top-1: 50.8\% vs. 49.5\%). Moreover, the ensemble of the deep model of 8 frame inputs and the shallow model of 16 frame inputs achieves the best accuracy (Top-1: 54.1\%). All these results further demonstrate the effectiveness and efficiency of the proposed method.

\section{Conclusion}
In this work, we presented the region-based non-local operation (RNL), a novel self-attention mechanism that effectively captures long-range dependencies by exploiting pair-wise region relationships. The RNL blocks can be easily embedded into the off-the-shelf CNNs architectures for end-to-end training. We have performed ablation studies to investigate the effectiveness of the proposed RNL operation in various settings. To verify the efficiency and effectiveness of the proposed methodology, we conducted experiments on two video benchmarks, Kinetics-400 and Something-Something V1. The results of the proposed method are shown to outperform the baseline and other recently proposed attention methods. Furthermore, we achieve state-of-the-art performance on Something-Something V1, which has demonstrated the powerful representation learning ability of our models.

\bibliographystyle{IEEEtran}
\bibliography{ms.bbl}

\begin{thebibliography}{10}
\providecommand{\url}[1]{#1}
\csname url@samestyle\endcsname
\providecommand{\newblock}{\relax}
\providecommand{\bibinfo}[2]{#2}
\providecommand{\BIBentrySTDinterwordspacing}{\spaceskip=0pt\relax}
\providecommand{\BIBentryALTinterwordstretchfactor}{4}
\providecommand{\BIBentryALTinterwordspacing}{\spaceskip=\fontdimen2\font plus
\BIBentryALTinterwordstretchfactor\fontdimen3\font minus
  \fontdimen4\font\relax}
\providecommand{\BIBforeignlanguage}[2]{{%
\expandafter\ifx\csname l@#1\endcsname\relax
\typeout{** WARNING: IEEEtran.bst: No hyphenation pattern has been}%
\typeout{** loaded for the language `#1'. Using the pattern for}%
\typeout{** the default language instead.}%
\else
\language=\csname l@#1\endcsname
\fi
#2}}
\providecommand{\BIBdecl}{\relax}
\BIBdecl

\bibitem{tran2015learning}
D.~Tran, L.~Bourdev, R.~Fergus, L.~Torresani, and M.~Paluri, ``Learning
  spatiotemporal features with {3D} convolutional networks,'' in \emph{Proc.
  IEEE Int. Conf. Comput. Vis. (ICCV)}, 2015, pp. 4489--4497.

\bibitem{carreira2017quo}
J.~Carreira and A.~Zisserman, ``Quo vadis, action recognition? a new model and
  the kinetics dataset,'' in \emph{Proc. IEEE Conf. Comput. Vis. Pattern Recog.
  (CVPR)}, 2017, pp. 4724--4733.

\bibitem{yue2015beyond}
J.~Yue-Hei~Ng, M.~Hausknecht, S.~Vijayanarasimhan, O.~Vinyals, R.~Monga, and
  G.~Toderici, ``Beyond short snippets: Deep networks for video
  classification,'' in \emph{Proc. IEEE Conf. Comput. Vis. Pattern Recog.
  (CVPR)}, 2015, pp. 4694--4702.

\bibitem{donahue2015long}
J.~Donahue, L.~Anne~Hendricks, S.~Guadarrama, M.~Rohrbach, S.~Venugopalan,
  K.~Saenko, and T.~Darrell, ``Long-term recurrent convolutional networks for
  visual recognition and description,'' in \emph{Proc. IEEE Conf. Comput. Vis.
  Pattern Recog. (CVPR)}, 2015, pp. 2625--2634.

\bibitem{he2016deep}
K.~He, X.~Zhang, S.~Ren, and J.~Sun, ``Deep residual learning for image
  recognition,'' in \emph{Proc. IEEE Conf. Comput. Vis. Pattern Recog. (CVPR)},
  2016, pp. 770--778.

\bibitem{hochreiter1997long}
S.~Hochreiter and J.~Schmidhuber, ``Long short-term memory,'' \emph{Neural
  Comput.}, vol.~9, no.~8, pp. 1735--1780, 1997.

\bibitem{wang2018non}
X.~Wang, R.~Girshick, A.~Gupta, and K.~He, ``Non-local neural networks,'' in
  \emph{Proc. IEEE Conf. Comput. Vis. Pattern Recog. (CVPR)}, 2018, pp.
  7794--7803.

\bibitem{vaswani2017attention}
A.~Vaswani, N.~Shazeer, N.~Parmar, J.~Uszkoreit, L.~Jones, A.~N. Gomez,
  {\L}.~Kaiser, and I.~Polosukhin, ``Attention is all you need,'' in
  \emph{Proc. Adv. Neural Inf. Process. Syst. (NIPS)}, 2017, pp. 5998--6008.

\bibitem{buades2005non}
A.~Buades, B.~Coll, and J.-M. Morel, ``A non-local algorithm for image
  denoising,'' in \emph{Proc. IEEE Conf. Comput. Vis. Pattern Recog. (CVPR)},
  vol.~2.\hskip 1em plus 0.5em minus 0.4em\relax IEEE, 2005, pp. 60--65.

\bibitem{goyal2017something}
R.~Goyal, S.~E. Kahou, V.~Michalski, J.~Materzynska, S.~Westphal, H.~Kim,
  V.~Haenel, I.~Fruend, P.~Yianilos, M.~Mueller-Freitag \emph{et~al.}, ``The"
  something something" video database for learning and evaluating visual common
  sense.'' in \emph{ICCV}, vol.~1, no.~4, 2017, p.~5.

\bibitem{krizhevsky2012imagenet}
A.~Krizhevsky, I.~Sutskever, and G.~E. Hinton, ``Imagenet classification with
  deep convolutional neural networks,'' in \emph{Proc. Adv. Neural Inf.
  Process. Syst. (NIPS)}, 2012, pp. 1097--1105.

\bibitem{Simonyan15}
K.~Simonyan and A.~Zisserman, ``Very deep convolutional networks for
  large-scale image recognition,'' in \emph{Int. Conf. Learn. Rep. (ICLR)},
  2015.

\bibitem{szegedy2015going}
C.~Szegedy, W.~Liu, Y.~Jia, P.~Sermanet, S.~Reed, D.~Anguelov, D.~Erhan,
  V.~Vanhoucke, and A.~Rabinovich, ``Going deeper with convolutions,'' in
  \emph{Proc. IEEE Conf. Comput. Vis. Pattern Recog. (CVPR)}, 2015, pp. 1--9.

\bibitem{szegedy2016rethinking}
C.~Szegedy, V.~Vanhoucke, S.~Ioffe, J.~Shlens, and Z.~Wojna, ``Rethinking the
  inception architecture for comp. vision,'' in \emph{Proc. IEEE Conf. Comput.
  Vis. Pattern Recog. (CVPR)}, 2016, pp. 2818--2826.

\bibitem{inception_v4}
C.~Szegedy, S.~Ioffe, V.~Vanhoucke, and A.~Alemi, ``Inception-v4,
  inception-resnet and the impact of residual connections on learning,''
  \emph{AAAI Conference on Artificial Intelligence}, 2016.

\bibitem{xie2017aggregated}
S.~Xie, R.~Girshick, P.~Doll{\'a}r, Z.~Tu, and K.~He, ``Aggregated residual
  transformations for deep neural networks,'' in \emph{Proc. IEEE Comp. Vision
  and Pattern Recog. (CVPR)}, 2017, pp. 5987--5995.

\bibitem{huang2017densely}
G.~Huang, Z.~Liu, L.~Van Der~Maaten, and K.~Q. Weinberger, ``Densely connected
  convolutional networks,'' in \emph{Proc. IEEE Conf. Comput. Vis. Pattern
  Recog. (CVPR)}, 2017, pp. 4700--4708.

\bibitem{tan2019efficientnet}
M.~Tan and Q.~V. Le, ``Efficientnet: Rethinking model scaling for convolutional
  neural networks,'' in \emph{Proc. Int. Conf. Mach. Learn.}, 2019.

\bibitem{simonyan2014two}
K.~Simonyan and A.~Zisserman, ``Two-stream convolutional networks for action
  recognition in videos,'' in \emph{Proc. Adv. Neural Inf. Process. Syst.
  (NIPS)}, 2014, pp. 568--576.

\bibitem{li2018videolstm}
Z.~Li, K.~Gavrilyuk, E.~Gavves, M.~Jain, and C.~G. Snoek, ``Videolstm
  convolves, attends and flows for action recognition,'' \emph{Computer Vision
  and Image Understanding}, vol. 166, pp. 41--50, 2018.

\bibitem{huang2020learning}
G.~Huang and A.~G. Bors, ``Learning spatio-temporal representations with
  temporal squeeze pooling,'' in \emph{Proc. IEEE Int. Conf. on Acoustics,
  Speech and Signal Proc. (ICASSP)}, 2020.

\bibitem{wang2016temporal}
L.~Wang, Y.~Xiong, Z.~Wang, Y.~Qiao, D.~Lin, X.~Tang, and L.~Van~Gool,
  ``Temporal segment networks: Towards good practices for deep action
  recognition,'' in \emph{Proc. Eur. Conf. Comput. Vis. (ECCV), vol LNCS 9912},
  2016, pp. 20--36.

\bibitem{taylor2010convolutional}
G.~W. Taylor, R.~Fergus, Y.~LeCun, and C.~Bregler, ``Convolutional learning of
  spatio-temporal features,'' in \emph{Proc. Eur. Conf. Comput. Vis. (ECCV)},
  2010, pp. 140--153.

\bibitem{qiu2017learning}
Z.~Qiu, T.~Yao, and T.~Mei, ``Learning spatio-temporal representation with
  pseudo-3d residual networks,'' in \emph{Proc. IEEE Int. Conf. Comput. Vis.
  (ICCV)}, 2017, pp. 5533--5541.

\bibitem{lin2019tsm}
J.~Lin, C.~Gan, and S.~Han, ``Tsm: Temporal shift module for efficient video
  understanding,'' in \emph{Proc. IEEE Int. Conf. Comput. Vis. (ICCV)}, 2019,
  pp. 7083--7093.

\bibitem{xie2018rethinking}
S.~Xie, C.~Sun, J.~Huang, Z.~Tu, and K.~Murphy, ``Rethinking spatiotemporal
  feature learning: Speed-accuracy trade-offs in video classification,'' in
  \emph{Proc. Eur. Conf. Comput. Vis. (ECCV), vol. LNCS 11219}, 2018, pp.
  305--321.

\bibitem{tran2019video}
D.~Tran, H.~Wang, L.~Torresani, and M.~Feiszli, ``Video classification with
  channel-separated convolutional networks,'' in \emph{Proc. IEEE Int. Conf.
  Comput. Vis. (ICCV)}, 2019, pp. 5552--5561.

\bibitem{feichtenhofer2020x3d}
C.~Feichtenhofer, ``X3d: Expanding architectures for efficient video
  recognition,'' in \emph{Proc. IEEE Conf. Comput. Vis. Pattern Recog. (CVPR)},
  2020, pp. 203--213.

\bibitem{bahdanau2015neural}
D.~Bahdanau, K.~Cho, and Y.~Bengio, ``Neural machine translation by jointly
  learning to align and translate,'' in \emph{Int. Conf. Learn. Representations
  (ICLR)}, 2015.

\bibitem{Hu_2018_CVPR}
J.~Hu, L.~Shen, and G.~Sun, ``Squeeze-and-excitation networks,'' in \emph{Proc.
  IEEE Conf. Comput. Vis. Pattern Recog. (CVPR)}, 2018, pp. 7132--7141.

\bibitem{wang2017residual}
F.~Wang, M.~Jiang, C.~Qian, S.~Yang, C.~Li, H.~Zhang, X.~Wang, and X.~Tang,
  ``Residual attention network for image classification,'' in \emph{Proc. IEEE
  Conf. Comput. Vis. Pattern Recog. (CVPR)}, 2017, pp. 3156--3164.

\bibitem{woo2018cbam}
S.~Woo, J.~Park, J.~Lee, and I.~So~Kweon, ``Cbam: Convolutional block attention
  module,'' in \emph{Proc. Eur. Conf. Comput. Vis. (ECCV)}, 2018, pp. 3--19.

\bibitem{cao2019gcnet}
Y.~Cao, J.~Xu, S.~Lin, F.~Wei, and H.~Hu, ``Gcnet: Non-local networks meet
  squeeze-excitation networks and beyond,'' in \emph{Proc. IEEE Int. Conf.
  Comput. Vis. Workshops (ICCV-w)}, 2019.

\bibitem{yue2018compact}
K.~Yue, M.~Sun, Y.~Yuan, F.~Zhou, E.~Ding, and F.~Xu, ``Compact generalized
  non-local network,'' in \emph{Proc. Adv. Neural Inf. Process. Syst. (NIPS)},
  2018, pp. 6510--6519.

\bibitem{sandler2018mobilenetv2}
M.~Sandler, A.~Howard, M.~Zhu, A.~Zhmoginov, and L.~Chen, ``Mobilenetv2:
  Inverted residuals and linear bottlenecks,'' in \emph{Proc. IEEE Conf.
  Comput. Vis. Pattern Recog. (CVPR)}, 2018, pp. 4510--4520.

\bibitem{szegedy2015bn}
S.~Ioffe and C.~Szegedy, ``Batch normalization: Accelerating deep network
  training by reducing internal covariate shift,'' in \emph{Proc. Int. Conf.
  Mach. Learn. (ICML) - Volume 37}, 2015, p. 448–456.

\bibitem{imagenet_cvpr09}
J.~Deng, W.~Dong, R.~Socher, L.-J. Li, K.~Li, and L.~Fei-Fei, ``Imagenet: A
  large-scale hierarchical image database,'' in \emph{Proc. IEEE Conf. Comput.
  Vis. Pattern Recog. (CVPR)}, 2009, pp. 248--255.

\bibitem{feichtenhofer2019slowfast}
C.~Feichtenhofer, H.~Fan, J.~Malik, and K.~He, ``Slowfast networks for video
  recognition,'' in \emph{Proc. IEEE Conf. Comput. Vis. Pattern Recog. (CVPR)},
  2019, pp. 6202--6211.

\bibitem{qiu2019learning}
Z.~Qiu, T.~Yao, C.-W. Ngo, X.~Tian, and T.~Mei, ``Learning spatio-temporal
  representation with local and global diffusion,'' in \emph{Proc. IEEE Conf.
  Comput. Vis. Pattern Recog. (CVPR)}, 2019, pp. 12\,056--12\,065.

\bibitem{wang2018videos}
X.~Wang and A.~Gupta, ``Videos as space-time region graphs,'' in \emph{Proc.
  Eur. Conf. Comput. Vis. (ECCV)}, 2018, pp. 399--417.

\bibitem{li2020smallbignet}
X.~Li, Y.~Wang, Z.~Zhou, and Y.~Qiao, ``Smallbignet: Integrating core and
  contextual views for video classification,'' in \emph{Proc. IEEE Conf.
  Comput. Vis. Pattern Recog. (CVPR)}, 2020, pp. 1092--1101.

\end{thebibliography}

\end{document}